\documentclass{sig-alternate-05-2015}
\usepackage{fixltx2e}
\usepackage{graphicx}
\usepackage[table]{xcolor}
\usepackage{graphicx}
\usepackage{caption}
\usepackage{tabularx}
\usepackage{ragged2e}
\usepackage{kantlipsum}
\usepackage{mathtools, cuted}
\usepackage{multirow}
\usepackage{flushend}

\begin{document}

\setcopyright{acmcopyright}
\doi{10.475/123_4}

\isbn{123-4567-24-567/08/06}

\conferenceinfo{FPGA '17}{February 22--24, 2017, Monterey, CA, USA}

\title{Accurate and Efficient Hyperbolic Tangent Activation Function on FPGA using the DCT Interpolation Filter}

\numberofauthors{1} 
%
\author{
\alignauthor
Ahmed M. Abdelsalam, J.M. Pierre Langlois and F. Cheriet\\
       \affaddr{Computer and Software Engineering Department}\\
       \affaddr{Polytechnique Montreal}\\
       \affaddr{Montreal, Canada}\\
       \email{\{ahmed.abdelsalam, pierre.langlois, farida.cheriet\}@polymtl.ca }
}

\maketitle

\begin{abstract}
Implementing an accurate and fast activation function with low cost is a crucial aspect to the implementation of Deep Neural Networks (DNNs) on FPGAs. We propose a high-accuracy approximation approach for the hyperbolic tangent activation function of artificial neurons in DNNs. It is based on the Discrete Cosine Transform Interpolation Filter (DCTIF).  The proposed architecture combines simple arithmetic operations on stored samples of the hyperbolic tangent function and on input data. The proposed DCTIF implementation achieves two orders of magnitude greater precision than previous work while using the same or fewer computational resources. Various combinations of DCTIF parameters can be chosen to tradeoff the accuracy and complexity of the hyperbolic tangent function. In one case, the proposed architecture approximates the hyperbolic tangent activation function with 10\textsuperscript{-5} maximum error while requiring only 1.52 Kbits memory and 57 LUTs of a Virtex-7 FPGA. We also discuss how the activation function accuracy affects the performance of DNNs in terms of their training and testing accuracies. We show that a high accuracy approximation can be necessary in order to maintain the same DNN training and testing performances realized by the exact function.              
\end{abstract}


\section{Introduction}
Deep Neural Networks (DNN) have been widely adopted in several applications such as object classification, pattern recognition and regression problems [1]. Although DNNs achieve high performance in many applications, this comes at the expense of a large number of arithmetic and memory access operations for both training and testing [2]. Therefore, DNN accelerators are highly desired [3]. FPGA-based DNN accelerators are favorable since FPGA platforms support high performance, configurability, low power consumption and quick development process [3]. On the other hand, implementing a DNN or a Convolutional Neural Network (CNN) on an FPGA is a challenging task since DNNs and CNNs require a large amount of resources [4], [5] and [6].

DNNs consist of a number of hidden layers that work in parallel, and each hidden layer has a number of Artificial Neurons (AN) [1]. Each neuron receives signals from other neurons and computes a weighted-sum of these inputs. Then, an activation function of the AN is applied on this weighted-sum. One of the main purposes of the activation function is to introduce non-linearity into the network.  The hyperbolic tangent is one of the most popular non-linear activation functions in DNNs [1].

Realizing a precise implementation of the hyperbolic tangent activation function in hardware entails a large number of additions and multiplications [7]. This implementation would badly increase the overall resources required for implementing a single AN and a fully parallel DNN. Therefore, approximations with different precisions and amount of resources are generally employed [7]. We propose a new high-accuracy approximation using the Discrete Cosine Transform Interpolation Filter (DCTIF) [8]. The proposed DCTIF approximation achieves higher accuracy than the existing approximations, and it needs fewer resources than other designs when a high precision approximation is required. We also study the effect of approximating the hyperbolic tangent activation function on the performance of training and testing DNNs.  
       
The rest of the paper is organized as follows: Different tanh approximations are reviewed in Section 2. The operation principle of the proposed DCTIF approximation is described in Section 3. In Section 4, an implementation of the proposed DCTIF approximation is detailed. Section 5 is dedicated to the experimental results and a comparison with other approximations and discussion.  Finally, Section 6 concludes the paper.

\begin{figure}
\centering
\includegraphics[height=0.65\linewidth, width=\linewidth]{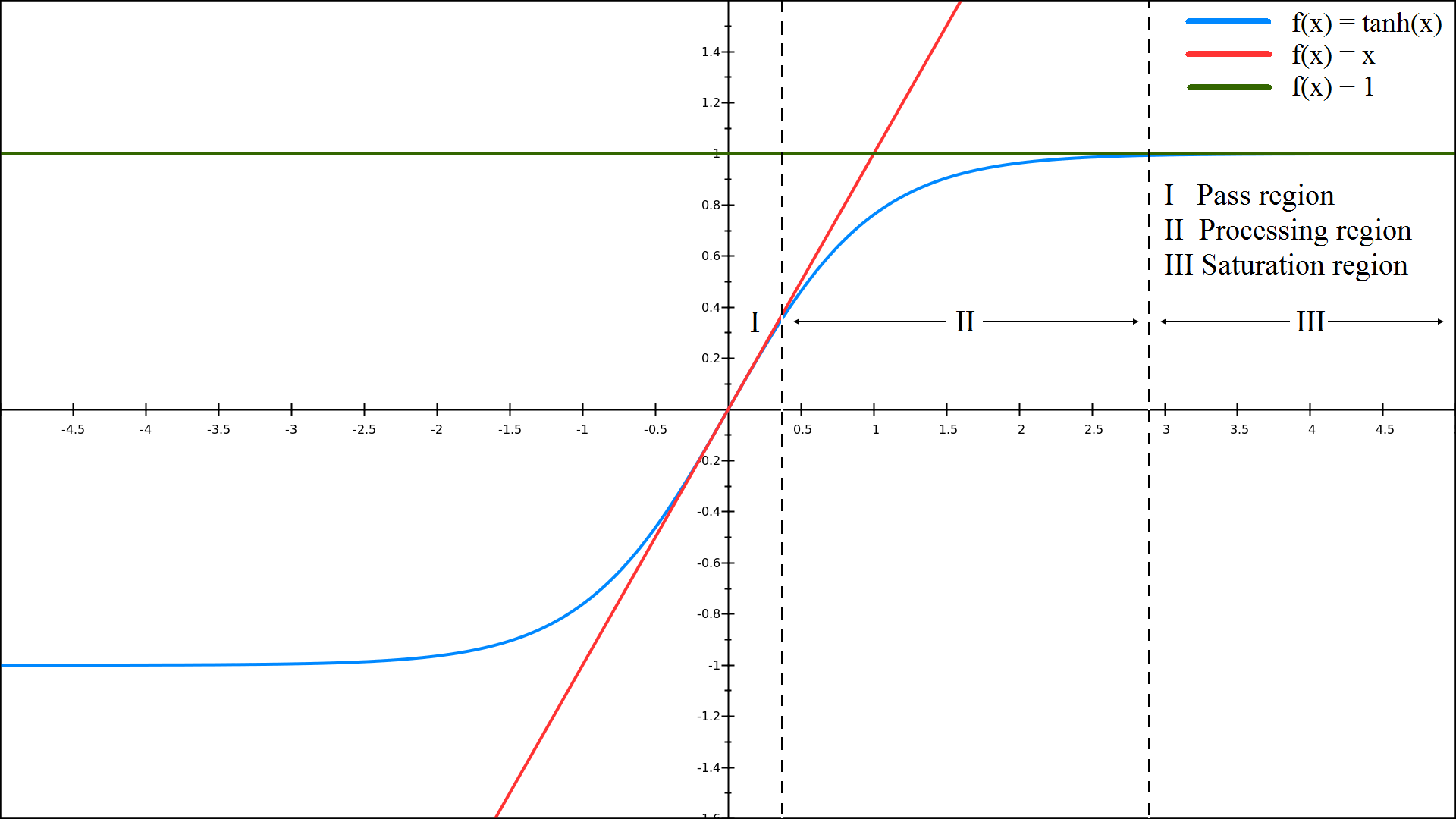}
\caption{The hyperbolic tangent activation function}
\end{figure}

\section{RELATED WORK}
The hardware implementation of a DNN is always constrained by the available computational resources [9]. The required computational resources to implement a DNN can be reduced by limiting the precision of the data representation [9]. On the other hand, using bitwise DNNs is another way to reduce the computational resources of a DNN. Bitwise DNN replaces floating or fixed-point arithmetic operations by efficient bitwise operations [10]. However, this comes at the expense of the training and testing performance of the DNN. Another approach to meet the constraints of the available computational resources is to approximate the activation function of the DNN. The selection of the tanh approximation accuracy as an activation function is one of the aspects that define the training and testing performance of the DNNs [11]. High accuracy approximations lead to high training and testing performance of the DNN, and low accuracy approximations lead to poor DNN performance [11].

There are several approaches for the hardware implementation of the hyperbolic tangent activation function based on Piecewise Linear (PWL), Piecewise Non-Linear, Lookup Table (LUT) and hybrid methods. All of these approaches exploit that the hyperbolic tangent function, shown in Figure 1, is negatively symmetric about the Y-axis. Therefore, the function can be evaluated for negative inputs by negating the output values of the same corresponding positive values and vice versa. Armato et al. [12] proposed to use PWL which divides the hyperbolic tangent function into segments and employs a linear approximation for each segment. On the other hand, Zhang and his colleagues [13] used a non-linear approximation for each segment. Although both methods achieve precise approximations for the hyperbolic tangent function, this comes at the expense of the throughput of the hardware implementation. LUT-based approximations divide the input range into sub-ranges where the output of each sub-range is stored in a LUT. Leboeuf et al. [14] proposed using a classical LUT and a Range Addressable LUT to approximate the function. LUT-based implementations are fast but they require more resources than PWL approximations in order to achieve the same accuracy. Therefore, most of the existing LUT-based methods limit the approximation accuracy to the range [0.02, 0.04]. 

Several authors noticed that the hyperbolic tangent function can be divided into three regions a) Pass Region, b) Processing Region (PR) and c) Saturation Region, as shown in Figure 1. The hyperbolic tangent function behaves almost like the identity function in the Pass Region, and its value is close to 1 in the Saturation Region. Some hybrid methods that combine LUTs and computations were used to approximate the non-linear PR. Namin and his colleagues [15] proposed to apply a PWL algorithm for the PR. On the other hand, Meher et al. [16] proposed to divide the input range of the PR into sub-ranges, and they implemented a decoder that takes the input value and selects which value should appear on the output port. Finally, Zamanloony et al. [7] introduced a mathematical analysis that defines the boundaries of the Pass, Processing and Saturation Regions of the hyperbolic tangent function based on the desired maximum error of the approximation.   

Generally, activation function approximations with high error badly affect the performance of DNNs in terms of their
training and testing accuracies. Approximations with higher accuracies are favorable in order to maintain the same learning capabilities and testing results compared to the exact activation function. Therefore, we propose a high precision approximation of the hyperbolic tangent activation function while using a small amount of computational resources.

\section{DCT INTERPOLATION FILTER DESIGN}
The DCT-based Interpolation Filter (DCTIF) interpolates data points from a number of samples of a function [6]. It was firstly introduced for interpolating fractional pixels from integer pixels in the motion compensation process of the latest video coding standard H.265 [6]. DCTIF can be used to approximate several non-linear functions. It interpolates values with a desired accuracy by controlling the number of samples involved in the interpolation process and the number of interpolated points between two samples. We propose to use DCTIF in order to approximate the hyperbolic activation function in DNNs.

The DCT transformation used to generate DCTIF coefficients is defined by Equation 1, where \textit{L\textsubscript{max}} and \textit{L\textsubscript{min}} define the range of the given sample points used in the interpolation process, \textit{Size} is defined as (\textit{L\textsubscript{max}} - \textit{L\textsubscript{min}} + \textit{1}) and the center position of a given size is \textit{Center} = (\textit{L\textsubscript{max}} + \textit{L\textsubscript{min}})/\textit{2}. By substituting Equation 1 into the inverse DCT formula defined in Equation 2, we get the DCTIF co-efficients generation formula for position \textit{i+r$\alpha$} as in Equation 3.

As shown in Figure 2, let's assume that \{\textit{p\textsubscript{2M}}\} denotes a set of \textit{2M} given sample points (no. of DCTIF filter's tabs) used to interpolate \textit{p\textsubscript{i+r$\alpha$}} at fractional position \textit{i+r$\alpha$} between two adjacent samples at positions \textit{i} and \textit{i+1} of the function \textit{x(n)}. The parameter \textit{$\alpha$} is a positive fractional number that is equal to (1/2\textsuperscript{j}) where \textit{j} is the number of interpolated points between two sample points. The parameter \textit{r} is a positive integer that represents the position of the interpolated point between two sample points where it is $\in$ [1, 2\textsuperscript{j}-1]. A fractional position value \textit{p\textsubscript{i+r$\alpha$}} is interpolated using an even number of samples when \textit{r$\alpha$} is equal to 1/2 , which means that the interpolated point is exactly between two adjacent samples. Otherwise, \textit{p\textsubscript{i+r$\alpha$}} is interpolated using an odd number of samples since the interpolated point is closer to one of the samples than the other. Therefore, Equation 3 is modified to generate the DCTIF co-efficients for even and odd numbers of tabs as in Equations 4 and 5, respectively. 

The DCTIF co-efficients can be smoothed using a smoothing window of size \textit{W} [8]. For hardware implementation, the smoothed co-efficients are scaled by a factor of (2\textsuperscript{s}) and rounded to integers, where \textit{s} is a positive integer value. In addition, the scaled co-efficients should be normalized which means that their summation is equal to 2\textsuperscript{s}. Consequently, Equation 6 defines the final DCTIF co-efficients.

\begin{strip}
\begin{equation}
X\left(k\right)=\sqrt{\frac{2}{Size}} \sum_{n=L_{min}}^{L_{max}}\left(x\left(n\right).\cos{\left(\frac{2n -\
\left(2\ \times{}\ Center\right)\ +\ Size}{2\ \times{}\
Size}\right)}\pi{}k\right)
\end{equation}

\begin{equation}
x\left(n\right)=\sqrt{\frac{2}{Size}} \sum_{k=L_{min}}^{L_{max}}\left(X\left(k\right).\cos{\left(\frac{2n -\
\left(2\ \times{}\ Center\right)\ +\ Size}{2\ \times{}\
Size}\right)}\pi{}k\right)
\end{equation}

\begin{equation}
x\left(i+r\alpha{}\right)=\frac{2}{Size} \sum_{k=L_{min}}^{L_{max}}\left(\cos{\left(\frac{2n -\
\left(2\ \times{}\ Center\right)\ +\ Size}{Size}\right)}\pi{}k\ .\
\cos{\left(\frac{2(i+r\alpha{}) -\ \left(2\ \times{}\ Center\right)\ +\
Size}{Size}\right)}\pi{}k\right)\
\end{equation}

\begin{equation}
{filter}_{even}(i+r\alpha{})=\frac{1}{M} \sum_{k=0}^{2M-1}\left(\cos{\left(\frac{2n -1+2M}{4M}\right)\pi{}k}\ .\ \cos{
⁡\left(\frac{2r\alpha{} -1+2M}{4M}\right)}\pi{}k\right)
\end{equation}

\begin{equation}
{filter}_{odd}(i+r\alpha{})=\frac{2}{2M+1} \sum_{k=0}^{2M}\left(\cos{\left(\frac{2n +1+2M}{2(2M+1)}\right)\pi{}k}\ .\ \cos{
⁡\left(\frac{2r\alpha{} +1+2M}{2(2M+1)}\right)}\pi{}k\right)
\end{equation}

\begin{equation}
{Filter}_{even/odd}(i+r\alpha{})={filter}_{even/odd}(i+r\alpha{}) .\ \cos{\left(\frac{n -r\alpha{}}{W-1}\right)\pi}\ .\ 2^s
\end{equation}
\end{strip}
Table 1 shows the generated DCTIF co-efficient values using different numbers of DCTIF tabs, \textit{r$\alpha$} values and scaling factors by substituting in Equation 6. The co-efficient values exihibit similarity among some \textit{r$\alpha$} positions. For example, the \textit{i+1/4} and \textit{i+3/4} positions have the same set of co-efficient values. Moreover, at the \textit{i+1/2} position, the set of co-efficients is symmetric about the center element. These properties can be exploited to reduce the implementation cost.

\begin{figure}
\centering
\includegraphics[height=0.8\linewidth, width=\linewidth]{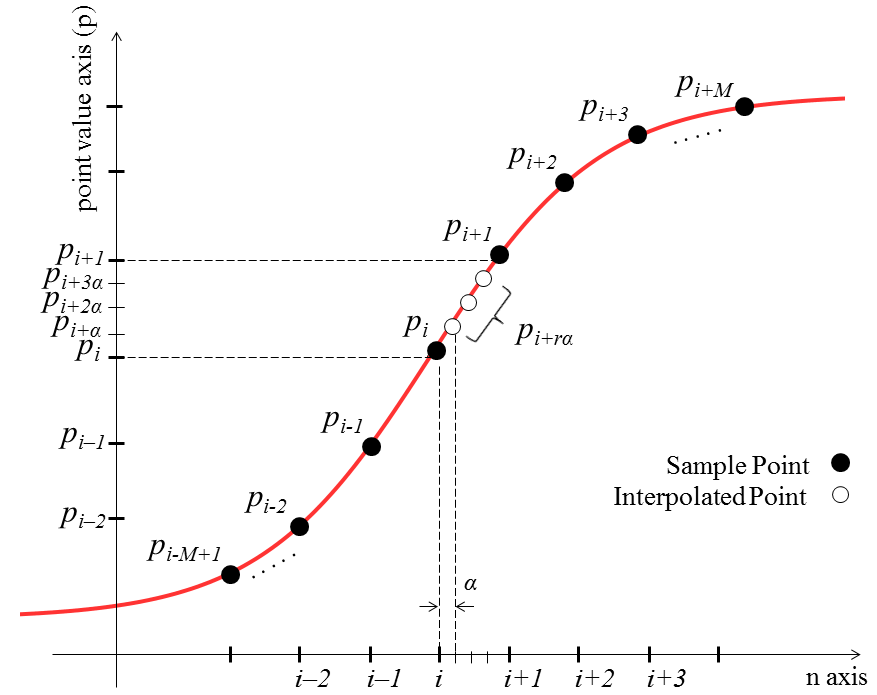}
\caption{DCT interpolation filter in tanh function approximation}
\end{figure}

A DCTIF approximation error analysis is presented in Figure 3. It can be seen that the DCTIF approximation error increases for small \textit{$\alpha$} values. Although a large \textit{$\alpha$} value means that fewer points need to be interpolated, this comes at the expense of memory resources since more samples must be stored. A large value of \textit{s} increases the accuracy of the approximation, but increases complexity as well because the interpolation coefficients take larger values, potentially expressed with more signed digits as shown in Table 1. Moreover, using more DCTIF tabs comes at the expense of the computational resources as shown in Table 2. 


\section{PROPOSED DCTIF ARCHITECTURE}
The proposed DCTIF approximation divides the input range of the hyperbolic tangent function into Pass, Processing and Saturation Regions as shown in Figure 1. The boundaries of these regions are computed based on the targeted maximum error of the approximation [7]. The output is equal to the input when the input is in the Pass Region. The proposed DCTIF approximation is utilized for the inputs in the Processing Region. In the Saturation Region, all the bits of the output port are set to one which represents the maximum value of the output signal. 

\begin{figure}
\centering
\includegraphics[height=0.7\linewidth, width=\linewidth]{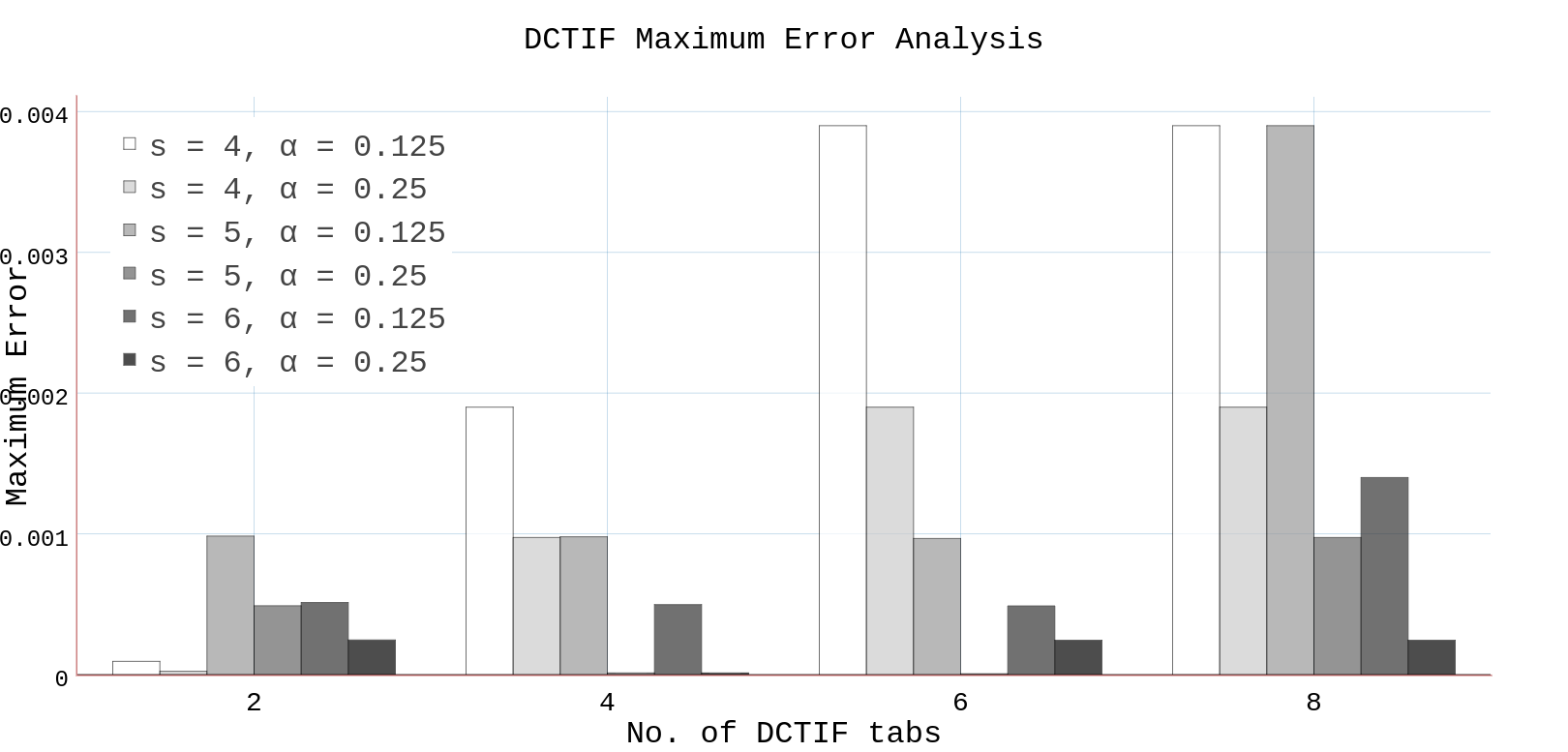}
\caption{DCTIF tanh approximation accuracy vs no. of tabs, \textit{$\alpha$} value and the scaling parameter \textit{s}}
\end{figure}

\begin{table*}[]
\centering
\caption{DCTIF co-efficient values for hyperbolic tangent approximation}
\label{table3}
\begin{tabular}{|c|c|c|c|}
\hline
No. of Tabs & Position (\textit{$\alpha$+ri}) & \begin{tabular}[c]{@{}c@{}}Filter Co-efficients\\ for \textit{s} = 4\end{tabular} & \begin{tabular}[c]{@{}c@{}}Filter Co-efficients \\ for \textit{s} = 5\end{tabular} \\ \hline
\multirow{3}{*}{4} & \textit{i+1/4} & \{-2, 15, 3, 0\} & \{-3, 29, 6, 0\} \\ \cline{2-4} 
 & \textit{i+1/2} & \{-2, 10, 10, -2\} & \{-3, 19, 19, -3\} \\ \cline{2-4} 
 & \textit{i+3/4} & \{0, 3, 15, -2\} & \{0, 6, 29, -3\} \\ \hline
\multirow{3}{*}{6} & \textit{i+1/4} & \{1, -2, 14, 4, -1, 0\} & \{1, -5, 29, 9, -2, 0\} \\ \cline{2-4} 
 & \textit{i+1/2} & \{1, -3, 10, 10, -3, 1\} & \{1, -5, 20, 20, -5, 1\} \\ \cline{2-4} 
 & \textit{i+3/4} & \{0, -1, 4, 14, -2, 1\} & \{0, -2, 9, 29, -5, 1\} \\ \hline
\end{tabular}
\end{table*}

\begin{table}[]
\centering
\caption{Complexity comparison of different DCTIF implementations on Xilinx FPGA Virtex-7 device}
\label{table4}
\begin{tabular}{|c|c|c|}
\hline
DCTIF Architecture & No. of LUTs & Delay (ns) \\ \hline
2-tabs, \textit{s} = 4, \textit{$\alpha$ = 1/4} & 21 & 1.640 \\ \hline
2-tabs, \textit{s} = 4, \textit{$\alpha$ = 1/8} & 37 & 2.023 \\ \hline
2-tabs, \textit{s} = 5, \textit{$\alpha$ = 1/8} & 41 & 2.439 \\ \hline
4-tabs, \textit{s} = 4, \textit{$\alpha$ = 1/4} & 50 & 5.588 \\ \hline
4-tabs, \textit{s} = 6, \textit{$\alpha$ = 1/4} & 57 & 7.432 \\ \hline
\end{tabular}
\end{table} 

The block diagram of the proposed architecture is shown in Figure 4. It is composed of a 4-input multiplexer that selects the appropriate output based on the input range decoder that determines the proper region of its input value. The decoder has four possible outputs that represent a) Pass Region, b) Saturation Region, c) Processing Region and the output is stored as a sample and finally d) Processing Region and the output of the given input needs to be interpolated. The truncation process shown in Figure 4 is implemented in order to pass the N\textsubscript{out} fraction bits of the input.

The implementation cost of the DCTIF approximation, shown in Figure 4, depends on the number of tabs and the values of \textit{s} and \textit{$\alpha$}. The cost of different DCTIF implementations are listed in Table 2 for five combinations of architectural parameter values. Figure 5 shows the DCTIF implementation using four tabs, \textit{s} = 4 and \textit{$\alpha$} = 1/4 where the co-efficient values are shown in Table 1. The interpolation equations are:

\begin{equation}
p_{i+1/4}=\ -2\ A+15\ B+3\ C-0\ D\
\end{equation}
\begin{equation}
p_{i+1/2}=\ -2\ A+10\ B+10\ C-2\ D\
\end{equation}
\begin{equation}
p_{i+3/4}=\ 0\ A+3\ B+15\ C-2\ D\
\end{equation}\

\begin{figure}
\centering
\includegraphics[height=0.6\linewidth, width=\linewidth]{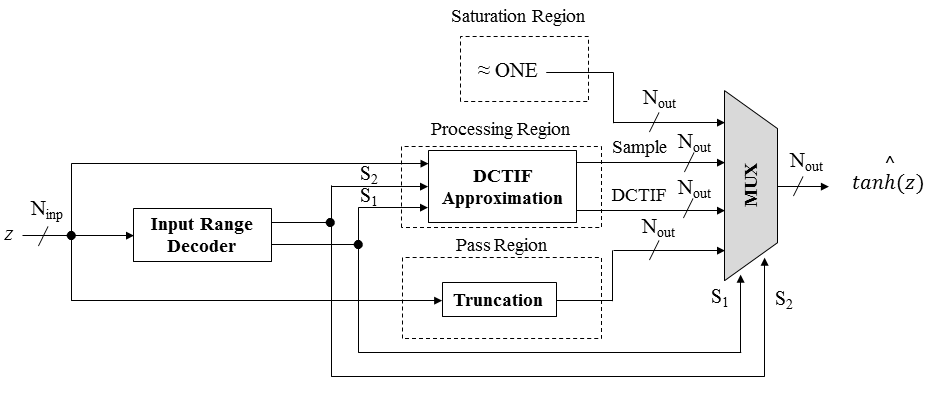}
\caption{Block diagram of the proposed tanh approximation using DCTIF}
\end{figure}

\begin{figure*}
\centering
\includegraphics[height=0.4\linewidth, width=0.8\textwidth]{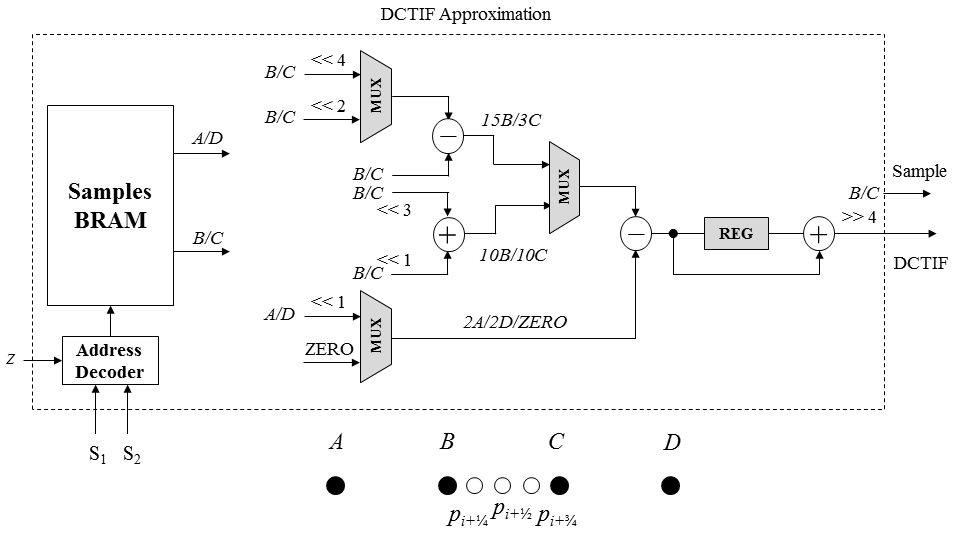}
\caption{The proposed DCTIF approximation architecture using 4 tabs, \textit{$\alpha$} = 1/4, \textit{s} = 4}
\end{figure*}

The address decoder of the DCTIF approximation, shown in Figure 5, takes the input value and the select lines of the input range decoder. It generates the addresses of the required samples (A, B, C, D) stored in the BRAM for the interpolation process. The samples A, B, C and D of Equations 7, 8 and 9 correspond to samples \textit{p\textsubscript{i-1}}, \textit{p\textsubscript{i}}, \textit{p\textsubscript{i+1}} and \textit{p\textsubscript{i+2}}, respectively, in Figure 2. Since the \textit{p\textsubscript{i+1/4}} and \textit{p\textsubscript{i+3/4}} interpolation equations are symmetric, the same hardware can be used to interpolate them. Therefore, we only implement the interpolation equations of \textit{p\textsubscript{i+1/4}} and \textit{p\textsubscript{i+1/2}}. In order to reduce the area required for the proposed implementation, we divide the computation of \textit{p\textsubscript{i+1/4}} and \textit{p\textsubscript{i+1/2}} equations into four pairs {(-2A + 15B), (3C + 0D), (-2A + 10B) and (10C - 2D)}. A set of three multiplexers, two subtractors and one adder, shown in Figure 5, is used to calculate the output value of any of these pairs. Each pair of these simple equations is computed in one clock cycle and the full equation takes two clock cycles to be calculated using an accumulator. A single cycle computation would also be possible, at the expense of more resources. Finally, the outputs of the DCTIF interpolation block are the interpolated value and the stored sample B when the input has its hyperbolic tangent output as a stored sample.

\section{EXPERIMENTAL RESULTS}
The proposed DCTIF approximation was described in Verilog HDL and synthesized for a Virtex-7 FPGA using Xilinx ISE 14.6. Table 3 compares the implemented DCTIF approximation to previous works in terms of maximum error, computational resources and throughput.

Table 3 shows that the proposed DCTIF approximation achieves 0.0002 maximum error while using only 21 Look-Up Tables (LUTs) and 1.12 kbits of memory. All existing works have been implemented as ASICs using TSMC 180 nm\textsuperscript{2} technology. The most accurate approximation achieves 0.01780 maximum error using 1,791 gates. The other works achieved the same approximation with less a amount of computational resources. Zamanloony and colleagues [7] achieved 0.01960 maximum error using only 129 gates. In addition, their implementation can be reconfigured in order to achieve higher accuracy at the expense of computational resources. In order to have a fair comparison, we re-implemented the design in [7] achieving 0.01180 maximum error for a Xilinx FPGA Virtex-7. We chose to re-implement the work in [7] as it requires the least amount of computational resources of all the existing implementations. Table 3 shows that our proposed DCTIF approximation outperforms the work in [7] in terms of accuracy using the same amount of resources. Therefore, we can say that the proposed DCTIF approximation outperforms the existing works in terms of accuracy using a similar amount of computational resources.

The proposed DCTIF approximation is based on interpolating the missing points in the Processing Region. High accuracy approximation can be achieved using the DCTIF approach by widening the boundaries of the Processing Region with respect to the two other regions. This directly increases the required amount of memory to store the sample values used in the interpolation process. In addition, more tabs of the interpolation filter must be used in order to achieve the target accuracy. This comes at the expense of the computational resources of the implementation as shown in Table 3. The proposed DCTIF approximation achieves 0.00001 maximum error, requiring only 1.52 kbits of memory and 57 LUTs. This implementation computes a value every 7.4 ns in two cycles of 3.2 ns each.

\begin{table*}[]
\centering
\caption{Complexity of different hyperbolic tangent approximations}
\label{my-label}
\begin{tabular}{|c|c|c|c|c|}
\hline
\multicolumn{5}{|c|}{ASIC Results on 180 nm\textsuperscript{2} TSMC Technology} \\ \hline
Architecture & Max. Error & Area (nm\textsuperscript{2}) & Gate Count & Delay (ns) \\ \hline
ICCIT {[}14{]} & 0.01800 & 17864.2 & 1791 & 2.45 \\ \hline
ICCIT {[}14{]} & 0.01780 & 11871.5 & 1190 & 2.12 \\ \hline
ISCAS {[}15{]} & 0.01890 & 5130.8 & 515 & 2.80 \\ \hline
VLSI-SOC {[}16{]} & 0.02050 & 1603.3 & 161 & 2.82 \\ \hline
TVLSI {[}7{]} & 0.01960 & 1280.3 & 129 & 2.12 \\ \hline
\multicolumn{5}{|c|}{FPGA Results on Xilinx Virtex-7} \\ \hline
Architecture & Max. Error & Slice LUTs & Memory (kbits) & Delay (ns) \\ \hline
TVLSI {[}7{]} & 0.01180 & 20 & --- & 1.245 \\ \hline
Proposed DCTIF \{2-tabs, \textit{s} = 4, and \textit{$\alpha$ = 1/4}\} & 0.00020 & 21 & 1.12 & 1.640 \\ \hline
Proposed DCTIF \{4-tabs, \textit{s} = 6, and \textit{$\alpha$ = 1/4}\} & 0.00001 & 57 & 1.52 & 7.432 \\ \hline
\end{tabular}
\end{table*}

The proposed DCTIF approximation can thus achieve high accuracy of the hyperbolic tangent activation function at low computational cost. In order to assess the impact of this accuracy of DNN performance, we trained and tested several DNN architectures. We conducted this experiment on two classification problems, MNIST [17] and CANCER [18], and Sinc and Sigmoid functions as regression problems [19].

Table 4 shows the testing performance of four different datasets with several DNN architectures while employing several approximations in the testing process. All the architectures in Table 4 were trained using the exact hyperbolic tangent activation function without any approximation. The Sinc and Sigmoid functions were sampled in the range [-3,3] with 600 samples each and used as regression problems [19]. Training and testing instances were selected randomly by 420 and 180 samples, respectively, for both functions. Sinc and Sigmoid functions results in Table 4 show that the normalized Mean Squared Error (MSE) value (MSE\textsubscript{approx} - MSE\textsubscript{exact}) is increased when using less accurate approximations for the same DNN architecture. In addition, the normalized MSE is getting larger when the DNN architecture becomes more complex with more number of hidden layers as shown in Figure 6. 

MNIST [17] and Cancer [18] are image classification data- sets. MNIST consists of 60,000 and 10,000 training and testing images, respectively, of the handwritten numbers 0 to 9. Cancer is a breast cancer dataset from UCI repository that has 699 images. MNIST results in Table 4 show that the testing accuracy of the classification process is highly affected by the precision of the approximation. Although the testing performance of Cancer dataset does not change with different approximations for the same DNN architecture, the normalized MSE is still increasing when using DNN architectures with large number of hidden layers as shown in Figure 6.

\begin{table*}[]
\centering
\caption{Testing errors of Sinc, Sigmoid, MNIST and Cancer datasets using different hyperbolic tangent approximations }
\label{my-label}
\begin{tabular}{|c|c|c|c|c|c|c|c|c|c|c|c|}
\hline
 & \begin{tabular}[c]{@{}c@{}}DNN \\ Architecture\end{tabular} & \begin{tabular}[c]{@{}c@{}}Tanh Max.\\ Error\end{tabular} & Correlation & \begin{tabular}[c]{@{}c@{}}Avg.\\ MSE\end{tabular} & \begin{tabular}[c]{@{}c@{}}Norm. \\ MSE\end{tabular} &  & \begin{tabular}[c]{@{}c@{}}DNN\\ Architecture\end{tabular} & \begin{tabular}[c]{@{}c@{}}Tanh Max.\\ Error\end{tabular} & \begin{tabular}[c]{@{}c@{}}Testing\\ Acc. (\%)\end{tabular} & \begin{tabular}[c]{@{}c@{}}Avg.\\ MSE\end{tabular} & \begin{tabular}[c]{@{}c@{}}Norm. \\ MSE\end{tabular} \\ \hline
\multirow{18}{*}{\rotatebox[origin=c]{90}{SINC}} & \multirow{6}{*}{\begin{tabular}[c]{@{}c@{}}4 Hidden\\ Layers x 5\\ ANs\end{tabular}} & 0.04 & 0.77237 & 0.1432 & 0.0608 & \multirow{18}{*}{\rotatebox[origin=c]{90}{MNIST}} & \multirow{6}{*}{\begin{tabular}[c]{@{}c@{}}1 Hidden \\ Layer x 30\\ ANs\end{tabular}} & 0.04 & 57.9 & 0.1960 & 0.0362 \\ \cline{3-6} \cline{9-12} 
 &  & 0.02 & 0.89024 & 0.1046 & 0.0216 &  &  & 0.02 & 72.8 & 0.1745 & 0.0147 \\ \cline{3-6} \cline{9-12} 
 &  & 0.01 & 0.92115 & 0.0902 & 0.0079 &  &  & 0.01 & 77.5 & 0.1657 & 0.0059 \\ \cline{3-6} \cline{9-12} 
 &  & 0.001 & 0.93335 & 0.0823 & 0.0002 &  &  & 0.001 & 77.8 & 0.1602 & 0.0004 \\ \cline{3-6} \cline{9-12} 
 &  & 0.0001 & 0.93373 & 0.0818 & 0.0000 &  &  & 0.0001 & 77.7 & 0.1599 & 0.0001 \\ \cline{3-6} \cline{9-12} 
 &  & 0 & 0.93376 & 0.0817 & 0 &  &  & 0 & 77.7 & 0.1598 & 0 \\ \cline{2-6} \cline{8-12} 
 & \multirow{6}{*}{\begin{tabular}[c]{@{}c@{}}6 Hidden\\ Layers x 5\\ ANs\end{tabular}} & 0.04 & 0.77797 & 0.1428 & 0.0615 &  & \multirow{6}{*}{\begin{tabular}[c]{@{}c@{}}10 Hidden\\ Layers x 30\\ ANs\end{tabular}} & 0.04 & 47.5 & 0.3045 & 0.1965 \\ \cline{3-6} \cline{9-12} 
 &  & 0.02 & 0.89177 & 0.1036 & 0.0229 &  &  & 0.02 & 73.4 & 0.1868 & 0.0788 \\ \cline{3-6} \cline{9-12} 
 &  & 0.01 & 0.91246 & 0.0899 & 0.0085 &  &  & 0.01 & 81.1 & 0.1369 & 0.0289 \\ \cline{3-6} \cline{9-12} 
 &  & 0.001 & 0.92553 & 0.0822 & 0.0006 &  &  & 0.001 & 82.6 & 0.1086 & 0.0006 \\ \cline{3-6} \cline{9-12} 
 &  & 0.0001 & 0.92601 & 0.0820 & 0.0001 &  &  & 0.0001 & 82.5 & 0.1080 & 0.0000 \\ \cline{3-6} \cline{9-12} 
 &  & 0 & 0.92605 & 0.0820 & 0 &  &  & 0 & 82.5 & 0.1080 & 0 \\ \cline{2-6} \cline{8-12} 
 & \multirow{6}{*}{\begin{tabular}[c]{@{}c@{}}8 Hidden\\ Layers x 5\\ ANs\end{tabular}} & 0.04 & 0.74849 & 0.1747 & 0.1018 &  & \multirow{6}{*}{\begin{tabular}[c]{@{}c@{}}20 Hidden\\ Layers x 30\\ ANs\end{tabular}} & 0.04 & 17.2 & 0.2881 & 0.1165 \\ \cline{3-6} \cline{9-12} 
 &  & 0.02 & 0.88751 & 0.1131 & 0.0402 &  &  & 0.02 & 22.3 & 0.2119 & 0.0403 \\ \cline{3-6} \cline{9-12} 
 &  & 0.01 & 0.92661 & 0.0874 & 0.0145 &  &  & 0.01 & 33.0 & 0.1867 & 0.0151 \\ \cline{3-6} \cline{9-12} 
 &  & 0.001 & 0.94100 & 0.0736 & 0.0007 &  &  & 0.001 & 38.1 & 0.1723 & 0.0007 \\ \cline{3-6} \cline{9-12} 
 &  & 0.0001 & 0.94137 & 0.0730 & 0.0001 &  &  & 0.0001 & 38.2 & 0.1716 & 0.0000 \\ \cline{3-6} \cline{9-12} 
 &  & 0 & 0.94140 & 0.0729 & 0 &  &  & 0 & 38.3 & 0.1716 & 0 \\ \hline
\multirow{18}{*}{\rotatebox[origin=c]{90}{Sigmoid}} & \multirow{6}{*}{\begin{tabular}[c]{@{}c@{}}4 Hidden\\ Layers x 5\\ ANs\end{tabular}} & 0.04 & 0.94904 & 0.2167 & 0.0050 & \multirow{18}{*}{\rotatebox[origin=c]{90}{Cancer}} & \multirow{6}{*}{\begin{tabular}[c]{@{}c@{}}1 Hidden\\ Layer x 3\\ ANs\end{tabular}} & 0.04 & 95.8 & 0.1166 & 0.0050 \\ \cline{3-6} \cline{9-12} 
 &  & 0.02 & 0.95131 & 0.2128 & 0.0011 &  &  & 0.02 & 95.8 & 0.1140 & 0.0024 \\ \cline{3-6} \cline{9-12} 
 &  & 0.01 & 0.95229 & 0.2122 & 0.0005 &  &  & 0.01 & 95.8 & 0.1128 & 0.0012 \\ \cline{3-6} \cline{9-12} 
 &  & 0.001 & 0.95308 & 0.2117 & 0.0000 &  &  & 0.001 & 95.8 & 0.1117 & 0.0001 \\ \cline{3-6} \cline{9-12} 
 &  & 0.0001 & 0.95315 & 0.2117 & 0.0000 &  &  & 0.0001 & 95.8 & 0.1116 & 0.0000 \\ \cline{3-6} \cline{9-12} 
 &  & 0 & 0.95316 & 0.2117 & 0 &  &  & 0 & 95.8 & 0.1116 & 0 \\ \cline{2-6} \cline{8-12} 
 & \multirow{6}{*}{\begin{tabular}[c]{@{}c@{}}6 Hidden\\ Layers x 5\\ ANs\end{tabular}} & 0.04 & 0.94403 & 0.2682 & 0.0601 &  & \multirow{6}{*}{\begin{tabular}[c]{@{}c@{}}10 Hidden\\ Layers x 3\\ ANs\end{tabular}} & 0.04 & 87.5 & 0.2663 & 0.0534 \\ \cline{3-6} \cline{9-12} 
 &  & 0.02 & 0.94615 & 0.2282 & 0.0201 &  &  & 0.02 & 87.5 & 0.2387 & 0.0258 \\ \cline{3-6} \cline{9-12} 
 &  & 0.01 & 0.95005 & 0.2160 & 0.0079 &  &  & 0.01 & 87.5 & 0.2255 & 0.0126 \\ \cline{3-6} \cline{9-12} 
 &  & 0.001 & 0.95268 & 0.2088 & 0.0007 &  &  & 0.001 & 87.5 & 0.2141 & 0.0012 \\ \cline{3-6} \cline{9-12} 
 &  & 0.0001 & 0.95304 & 0.2082 & 0.0001 &  &  & 0.0001 & 87.5 & 0.2130 & 0.0001 \\ \cline{3-6} \cline{9-12} 
 &  & 0 & 0.95305 & 0.2081 & 0 &  &  & 0 & 87.5 & 0.2129 & 0 \\ \cline{2-6} \cline{8-12} 
 & \multirow{6}{*}{\begin{tabular}[c]{@{}c@{}}8 Hidden\\ Layers x 5\\ ANs\end{tabular}} & 0.04 & 0.94042 & 0.4569 & 0.2394 &  & \multirow{6}{*}{\begin{tabular}[c]{@{}c@{}}20 Hidden\\ Layers x 3\\ ANs\end{tabular}} & 0.04 & 62.5 & 0.4314 & 0.2270 \\ \cline{3-6} \cline{9-12} 
 &  & 0.02 & 0.94836 & 0.2821 & 0.0646 &  &  & 0.02 & 62.5 & 0.3828 & 0.1784 \\ \cline{3-6} \cline{9-12} 
 &  & 0.01 & 0.95220 & 0.2326 & 0.0151 &  &  & 0.01 & 91.7 & 0.2139 & 0.0095 \\ \cline{3-6} \cline{9-12} 
 &  & 0.001 & 0.95435 & 0.2182 & 0.0007 &  &  & 0.001 & 91.7 & 0.2050 & 0.0006 \\ \cline{3-6} \cline{9-12} 
 &  & 0.0001 & 0.95447 & 0.2176 & 0.0001 &  &  & 0.0001 & 91.7 & 0.2045 & 0.0001 \\ \cline{3-6} \cline{9-12} 
 &  & 0 & 0.95448 & 0.2175 & 0 &  &  & 0 & 91.7 & 0.2044 & 0 \\ \hline
\end{tabular}
\end{table*}

Table 5 shows the training accuracy of the four datasets employing the hyperbolic tangent activation function with five approximations and the exact hyperbolic tangnet function in the training process of the network. The training accuracies of classification and regression problems decrease even when using precise hyperbolic tangent approximations with a maximum error of 10\textsuperscript{-4}. We noticed that when the networks are trained using less accurate approximations, the training process stops early before applying the full number of epochs. Therefore, the training accuracies are badly affected compared to the training accuracies using the exact hyperbolic tangent activation function. Moreover, that would degrade the overall testing results of both classification and regression problems.

\begin{figure*}
\centering
\includegraphics[height=0.55\linewidth, width=0.95\textwidth]{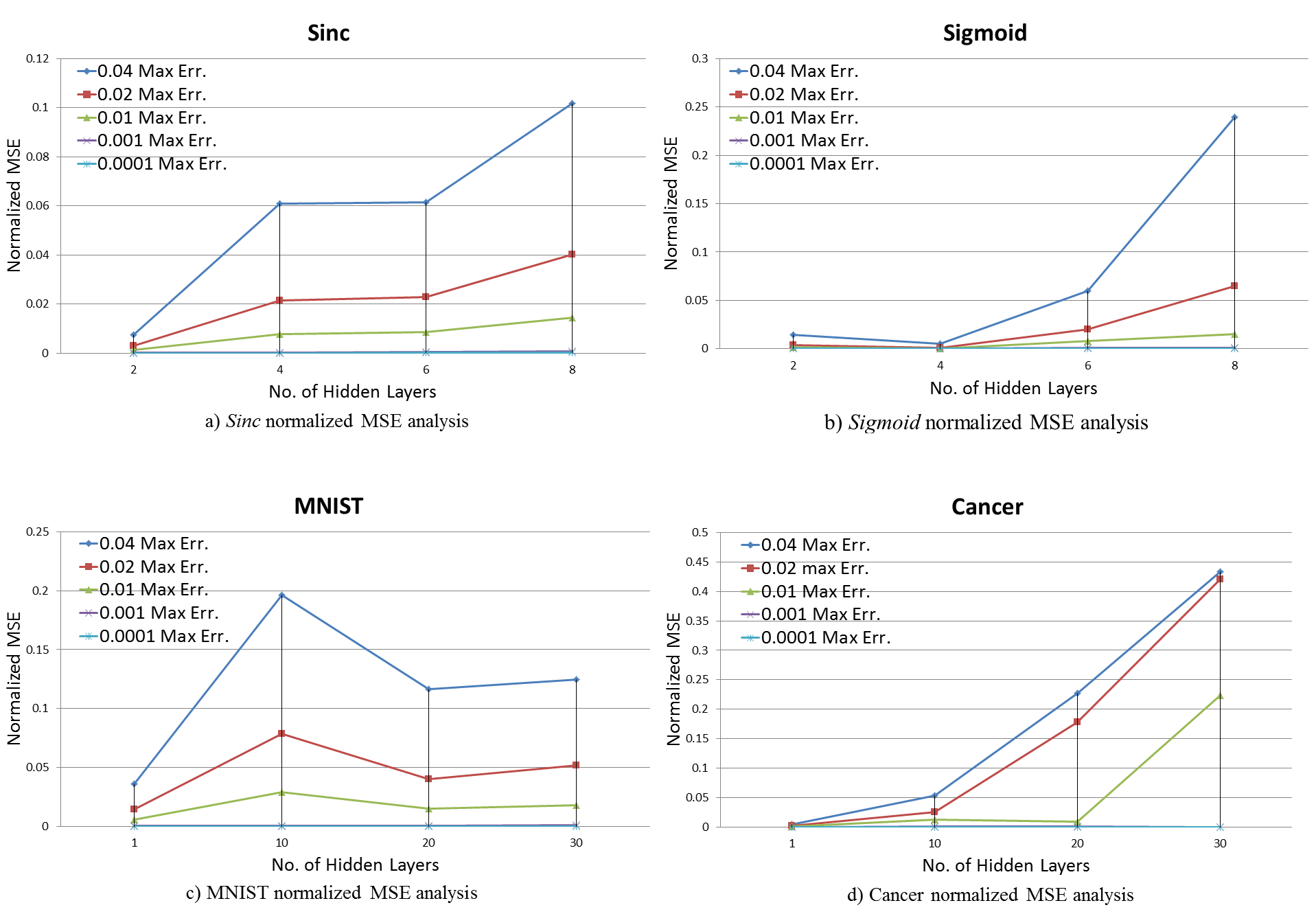}
\caption{Performance analysis of testing different DNNs architectures employing hyperbolic tangent activation function with different accuracies}
\end{figure*}

Generally, we showed that the performance of some widely used DNN architectures change using five hyperbolic tangent approximations with different accuracies. In some cases, a hyperbolic tangent function approximation with 10\textsuperscript{-5} is required in order to achieve the same performance of the exact function. Although implementing an approximation with high accuracy improves DNN performance, this requires more computational and memory resources and reduces the implementation's throughput.  The proposed DCTIF approach achieves such an accurate approximation while using small amount of computational and memory resources.      

\section{Conclusions}
The accuracy of the activation function is a bottleneck of the performance DNNs' implementations on FPGA. We studied how the accuracy of the hyperbolic tangent activation function approximation changes the performance of different DNNs. We proposed a high-accuracy approximation technique that is based on Discrete Cosine Transform Interpolation Filter. The proposed DCTIF approach outperforms the existing works in terms of accuracy for similar amounts of computational resources. Moreover, it achieves better approximation 
accuracy at the expense of computational and memory resources. We showed specific cases of DNN classification and regression problems where the high accuracy afforded by our approach results in significantly better training and testing performances.

\begin{table*}[]
\centering
\caption{Training errors of Sinc, Sigmoid, MNIST and Cancer using different hyperbolic tangent approximations }
\label{my-label}
\begin{tabular}{|c|c|c|c|c|c|}
\hline
\begin{tabular}[c]{@{}c@{}}DNN \\ Architecture\end{tabular} & \begin{tabular}[c]{@{}c@{}}Tanh Max. \\ Error\end{tabular} & Correlation & \begin{tabular}[c]{@{}c@{}}DNN\\ Architecture\end{tabular} & \begin{tabular}[c]{@{}c@{}}Tanh Max. \\ Error\end{tabular} & \begin{tabular}[c]{@{}c@{}}Training \\ Acc. (\%)\end{tabular} \\ \hline
 & 0.04 & 0.43279 &  & 0.04 & 10.7 \\ \cline{2-3} \cline{5-6} 
 & 0.02 & 0.78250 &  & 0.02 & 16.4 \\ \cline{2-3} \cline{5-6} 
 & 0.01 & 0.78976 &  & 0.01 & 23.1 \\ \cline{2-3} \cline{5-6} 
 & 0.001 & 0.84850 &  & 0.001 & 31.1 \\ \cline{2-3} \cline{5-6} 
 & 0.0001 & 0.87712 &  & 0.0001 & 68.0 \\ \cline{2-3} \cline{5-6} 
\multirow{-6}{*}{\begin{tabular}[c]{@{}c@{}}Sinc \\ 8 Hidden\\  Layers x 5 \\ ANs, 10,000 \\ epoch\end{tabular}} & 0 & 0.90287 & \multirow{-6}{*}{\begin{tabular}[c]{@{}c@{}}MNIST\\ 1 Hidden \\ Layer x 15 \\ ANs, 10,000\\ epoch\end{tabular}} & 0 & 68.1 \\ \hline
 & 0.04 & 0.77945 &  & 0.04 & 86.1 \\ \cline{2-3} \cline{5-6} 
 & 0.02 & 0.80033 &  & 0.02 & 86.9 \\ \cline{2-3} \cline{5-6} 
 & 0.01 & 0.80068 &  & 0.01 & 86.9 \\ \cline{2-3} \cline{5-6} 
 & 0.001 & 0.84581 &  & 0.001 & 86.9 \\ \cline{2-3} \cline{5-6} 
 & 0.0001 & 0.85014 &  & 0.0001 & 94..1 \\ \cline{2-3} \cline{5-6} 
\multirow{-6}{*}{\begin{tabular}[c]{@{}c@{}}Sigmoid \\ 8 Hidden \\ Layers x 5 \\ ANs, 10,000 \\ epoch\end{tabular}} & 0 & 0.86097 & \multirow{-6}{*}{\begin{tabular}[c]{@{}c@{}}Cancer\\ 1 Hidden \\ Layer x 15 \\ ANs, 10,000\\ epoch\end{tabular}} & 0 & 94.1 \\ \hline
\end{tabular}
\end{table*}

\section{Acknowledgments}
The authors would like to thank Ahmed El-Sheikh, Awny M. El-Mohandes and Hamza Bendaoudi for their insightful comments on our work.

\end{document}